\documentclass[runningheads]{llncs}
\usepackage{enumitem}
\usepackage{multicol}
\usepackage{graphicx}
\usepackage{times}
\usepackage{soul}
\usepackage{subcaption}
\usepackage{url}
\usepackage[hidelinks]{hyperref}
\usepackage[utf8]{inputenc}
\usepackage{caption}
\usepackage{graphicx}
\usepackage{amsmath}
\usepackage{amssymb}
\usepackage{booktabs}
\usepackage[linesnumbered,ruled,vlined]{algorithm2e}
\usepackage{algpseudocode}
\usepackage{graphicx}
\usepackage{tikz}
\usetikzlibrary{backgrounds}
\usepackage{tikz-qtree}
\usetikzlibrary{shapes, arrows}
\usepackage{float}
\usepackage{subcaption}
\usepackage{caption}
\usepackage{array}
{\setlength{\intextsep}{10pt}
\newcolumntype{x}[1]{>{\centering\arraybackslash}p{#1}}
\begin{document}
\title{Towards a Generic Representation of Combinatorial Problems for Learning-Based
Approaches}
%

%
\author{
Léo Boisvert\inst{1}\and Hélène Verhaeghe\inst{2}\orcidID{0000-0003-0233-4656} 
\and \\ Quentin Cappart\inst{1}\orcidID{0000-0002-8742-0774}
}

\institute{
Polytechnique Montréal, Montreal, Canada\\
\email{ \{leo.boisvert, quentin.cappart\}@polymtl.ca} \\\and
DTAI, KULeuven, Belgium \\
\email{helene.verhaeghe@kuleuven.be}
}
\authorrunning{L. Boisvert et al.}
\titlerunning{Towards a Generic Representation of Combinatorial Problems}
%
\maketitle              
\begin{abstract}
In recent years, there has been a growing interest in using learning-based approaches for solving combinatorial problems, either in an end-to-end manner or in conjunction with traditional optimization algorithms. In both scenarios, the challenge lies in encoding the targeted combinatorial problems into a structure compatible with the learning algorithm. Many existing works have proposed problem-specific representations, often in the form of a graph, to leverage the advantages of  \textit{graph neural networks}.
However, these approaches lack generality, as the representation cannot be easily transferred from one combinatorial problem to another one. While some attempts have been made to bridge this gap, they still offer a partial generality only. In response to this challenge, this paper advocates for progress toward a fully generic representation of combinatorial problems for learning-based approaches. The approach we propose involves constructing a graph by breaking down any constraint of a combinatorial problem into an abstract syntax tree and expressing relationships (e.g., a variable involved in a constraint) through the edges. Furthermore, we introduce a graph neural network architecture capable of efficiently learning from this representation. The tool provided operates on combinatorial problems expressed in the XCSP3 format, handling all the constraints available in the 2023 mini-track competition. Experimental results on four combinatorial problems demonstrate that our architecture achieves performance comparable to dedicated architectures while maintaining generality. Our code and trained models are publicly available at \url{https://github.com/corail-research/learning-generic-csp}.

\end{abstract}

\section{Introduction}

{\it Combinatorial optimization} has drawn the attention of computer scientists since the discipline emerged. Combinatorial problems, such as the {\it traveling salesperson problem} or the {\it Boolean satisfiability} have been the focus of many decades of research in the computer science community. We can now solve large problems of these kinds efficiently with exact methods~\cite{Applegate2009} and heuristics \cite{36b9628e7a874d208624584d8a470985}. Heuristics are handcrafted procedures that crystallize expert knowledge and intuition about the structure of the problems they are applied to. While heuristics have found success and applications for the resolution of combinatorial problems either as a direct-solving process or integrated into a search procedure, the  rise of deep learning in many different fields \cite{Bahdanau_Cho_Bengio_2016,brown2020language,Krizhevsky_Sutskever_Hinton_2012,mnih2015human} has attracted the attention of researchers~\cite{Prouvost_2020}. Among deep learning architectures,
\textit{graph neural networks} (GNNs)~\cite{Scarselli_Gori_Tsoi_Hagenbuchner_Monfardini_2009} have proven to be a powerful and flexible tool for solving combinatorial problems.
However, as identified by Cappart et al.~(2023)~\cite{cappart2021combinatorial},  
it is still cumbersome to integrate GNN and machine learning into existing solving processes
for practitioners. One reason is that a dedicated architecture must be designed and trained for each
combinatorial problem. In addition to potentially expensive computing resources for training, 
this also requires the existence of a large and labeled training set.

Building such a problem-specific graph representation has been the preferred choice
of most related approaches, such as NeuroSAT~\cite{Selsam2018} leveraging an encoding for SAT
formulas, or approaches tightly linked to the
traveling salesperson problem~\cite{Prates_Avelar_Lemos_Lamb_Vardi_2018,joshi2022learning}.
These approaches suffer from a lack of generality as the architecture could not be trivially exported from one combinatorial problem to another one (e.g., the representation in NeuroSAT cannot be used for encoding an instance of 
the traveling salesperson problem). 
Few attempts have been realized to bridge this gap, but they still provide only
a partial genericity. For instance, Gasse et al.~(2019)~\cite{Gasse_Chetelat_Ferroni_Charlin_Lodi_2019}
proposed a bipartite graph linking variables and constraints 
when a variable is involved in a given constraint. 
However, this approach only encodes binary mixed-integer programs. 
Chalumeau et al.~(2021)~\cite{chalumeau2021seapearl} later introduced a tripartite graph 
where variables, values, and constraints are specific types of vertices. 
This approach also lacks genericity as the method requires retraining when the number of variables changes.
To our knowledge, the last attempt in this direction has been realized 
by Marty et al.~(2023)~\cite{Marty_François_Tessier_Gauthier_Rousseau_Cappart_2023} who leveraged a tripartite graph allowing decorating each vertex type with dedicated features. Although any combinatorial problem can theoretically be encoded with this framework, some information is lost with the encoding.
For instance, it was not clear how the constraint $3x_1  \leq 4x_2$, could be differentiated from the constraint $2x_1 \leq 5x_2$. On the one hand, both constraints can be represented as an \textit{inequality}, but we lose information
about the variables' coefficients. On the other hand,  both constraints can be encoded as two distinct relationships, but
in this case, we lose the fact that we have an inequality in both cases. More formally, their encoding function was not \textit{injective}. Different instances could have the same encoding with no option  to differentiate them. Besides, the experiments proposed only targeted relatively pure problems (\textit{maximum independent set}, \textit{maximum cut}, and \textit{graph coloring}). 
Similar limitations are observed in the approach of Tönshoff et al.~(2022)~\cite{Tönshoff_Kisin_Lindner_Grohe_2022}.

Based on this context, this paper progresses towards a fully generic representation of combinatorial
problems for learning-based approaches. Intuitively, our idea is to break down any constraint of a problem instance as an \textit{abstract syntax tree} and connect similar items (e.g., the same variables or constraints)
through an edge. Then, we introduce a GNN architecture able to leverage this graph. 
To demonstrate the  genericity of this approach, our architecture directly operates on instances expressed with the \textsc{XCSP3} format~\cite{Boussemart_Lecoutre_Audemard_Piette_2022}
and can handle all the constraints available in the 2023 mini-track competition. Experiments are carried out on four problems (featuring standard intension, and global constraints such as \textsc{allDifferent}~\cite{regin1994filtering}, 
\textsc{table}~\cite{demeulenaere2016compact}, \textsc{negativeTable}~\cite{verhaeghe2017extending}, \textsc{element} and \textsc{sum}) and aims to predict the satisfiability of the decision version of combinatorial problems. 
The results show that our generic architecture gives performances close to problem-specific architectures and
outperforms the tripartite graph of Marty et al.~(2023)~\cite{Marty_François_Tessier_Gauthier_Rousseau_Cappart_2023}.


\section{Encoding Combinatorial Problem Instances as a Graph}

\begin{figure}
\begin{subfigure}[b]{0.5\textwidth}
\begin{align}
\mathsf{max} ~ &  x_1 &  \notag \\
\mathsf{s. t.} ~ &  3x_1 \leq 4x_2  &   \notag \\
& \textsc{table}([x_1,x_2],[(1,2),(2,3)]) &  \notag   \\
&  x_1 \in \{1,2 \} &   \notag \\
&  x_2 \in \{2,3 \} &   \notag 
\end{align}
\end{subfigure}
\begin{subfigure}[b]{1\textwidth}
\resizebox{0.4\textwidth}{!}{
        \begin{tikzpicture}[>=stealth',auto,node distance=2.0cm,
          thick,main node/.style={circle,fill=blue!20,draw,
          font=\sffamily\Large\bfseries,minimum size=10mm}]
          \pgfdeclarelayer{background}
          \pgfsetlayers{background,main}

          \node[main node, fill=green!25] (value_1) at (-4, 2) {$1$};
          \node[main node, fill=green!25] (value_2) at (-0, 2) {$2$};
          \node[main node, fill=green!25] (value_3) at (4, 2) {$3$};
          
          \node[main node, fill=red!90] (x_1) at (-2.5,-1) {$x_1$};
          \node[main node, fill=red!90] (x_2) at (2.5,-1) {$x_2$};

          \node[main node, fill=orange!50] (mul_1) at (-2.5, 0.5) {$\times 3$};
          \node[main node, fill=orange!50] (mul_2) at (2.5, 0.5) {$\times 4$};

          \node[main node, fill=orange!50] (lhs) at (-2.5, 2) {$lhs$};
          \node[main node, fill=orange!50] (rhs) at (2.5, 2) {$rhs$};

          \node[main node, fill=blue!40] (leq) at (0,0.5) {$\leq$};
          
          \node[main node, fill=orange!50] (tup1) at (-3,-3)  {$t_1$};
          \node[main node, fill=orange!50] (tup2) at (3,-3)  {$t_2$};
          
          \node[main node, fill=orange!50, minimum size=5mm] (combo11) at (-4,-2)  {$.$};
          \node[main node, fill=orange!50, minimum size=5mm] (combo22) at (-2,-2)  {$.$};

          \node[main node, fill=orange!50, minimum size=5mm] (combo21) at (2,-2)  {$.$};
          \node[main node, fill=orange!50, minimum size=5mm] (combo23) at (4, -2)  {$.$};
      
          \node[main node, fill=blue!40] (ext) at (0,-3) {$ext$};
          \node[main node, fill=yellow!40] (model) at (0,-1) {$M$};
          \begin{pgfonlayer}{background}
            \draw[rounded corners, fill=gray!40, opacity=0.5] 
                ([shift={(-0.25cm,-0.25cm)}]x_1.south west) rectangle 
                ([shift={(0.25cm,0.25cm)}]lhs.north east);
            \draw[rounded corners, fill=gray!40, opacity=0.5] 
                ([shift={(-0.25cm,-0.25cm)}]x_2.south west) rectangle 
                ([shift={(0.25cm,0.25cm)}]rhs.north east);
            \draw[rounded corners, fill=gray!40, opacity=0.5] 
                ([shift={(-0.25cm,-0.25cm)}]leq.south west) rectangle 
                ([shift={(0.25cm,0.25cm)}]leq.north east);
          \end{pgfonlayer}

          \path[every node/.style={font=\sffamily\small, fill=white,}]
            (x_1) edge[black] (value_1)
            (x_1) edge[black] (value_2)
            (x_1) edge[black] (mul_1)
            (mul_1) edge[black] (lhs)
            (rhs) edge[black] (leq)
            
            (x_2) edge[black] (value_2)
            (x_2) edge[black] (value_3)
            (x_2) edge[black] (mul_2)
            (mul_2) edge[black] (rhs)
            (lhs) edge[black] (leq)
            
            (x_1) edge[black] (combo21)
            (x_1) edge[black] (combo11)

            (x_2) edge[black] (combo22)
            (x_2) edge[black] (combo23)

            (value_1) edge[black] (combo11)
            (value_2) edge[black] (combo22)
            (value_2) edge[black] (combo21)
            (value_3) edge[black] (combo23)

            (tup1) edge[black] (combo11)
            (tup1) edge[black] (combo22)

            (tup2) edge[black] (combo21)
            (tup2) edge[black] (combo23)
                            
            (tup1) edge[black] (ext)
            (tup2) edge[black] (ext)
            
            (model) edge[black] (ext)
            (model) edge[black] (leq)
            (model) edge[black] (x_1)
            ;
        \end{tikzpicture}
}
\end{subfigure}
\caption{Encoding of a combinatorial problem instance presented as a running example.  There are $3$ \textit{value-vertices} depicted (in green) and $2$ \textit{variable-vertices} (in red).
As $x_1$ contains values $1$ and $2$ in its domain, they are connected with an edge, and similarly for the domain of $x_2$.
There are $2$ \textit{constraint-vertices} (in blue), one
for the inequality ($\leq$) and one for the \textsc{table} constraint ($\textsf{ext}$). The figure's gray area illustrates the constraint $3x_1 \leq 4x_2$, highlighted by operators (in orange) and showing a multiplication ($\times$, with feature $3$) of $x_1$ on the right-side ($\textsf{rhs}$) and another ($\times$, with feature $4$) of $x_2$ on the left-side ($\textsf{lhs}$). The $\textsf{rhs}$ and $\textsf{lhs}$ operators clarify equation sides, essential for distinguishing between $3x_1 \leq 4x_2$ and $3x_1 \geq 4x_2$, and link to the associated constraint (e.g., inequality $\leq$).
The constraint $\textsc{table}([x_1,x_2],[(1,2),(2,3)])$ is expressed in a similar way. It involves two tuples $t_1$ and $t_2$. Finally,  the \textit{model-vertex} (in yellow) is connected to the two constraints, and to variable $x_1$, as it is part of the objective function.}
   \label{fig:encoding}
\end{figure}
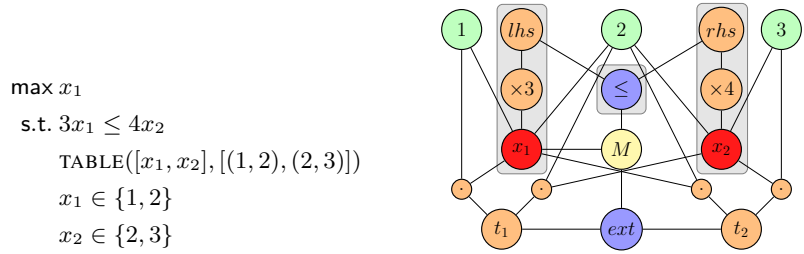

Formally, a combinatorial problem instance $\mathcal{P}$ is defined as a tuple $\langle X, D(X), C, O \rangle$, where $X$ is the set
of variables, $D(X)$ is the set of domains, $C$ is the set of 
constraints, and $O$ ($X \to \mathbb{R}$) is an objective function. 
A valid solution is an assignment of each variable to a value of its domain such that every constraint is satisfied. The optimal solution is a valid solution such that no other solution has a better value for the objective.
Our goal is to build a function $\Phi: \langle X, D(X), C, O \rangle \mapsto \mathcal{G}(V,f,E)$, where $\mathcal{G}$ is a  graph and $V$, $f$, $E$ are
its vertex, vertex features and arc sets, respectively.
We want this function to be \textit{injective}, i.e., an encoding refers to at most one combinatorial problem instance.
We propose to do so by introducing an encoding consisting in a heterogeneous and undirected graph featuring 5 types of vertices:
\textit{variables} ($\textsc{var}$), \textit{constraints} ($\textsc{cst}$), \textit{values} ($\textsc{val}$), 
\textit{operators} ($\textsc{ope}$), and \textit{model} ($\textsc{mod}$).
The idea is to split each constraint as
a sequence of elementary operations, to merge vertices representing
the same variable or value, and connect together all the relationships.
An illustration of this encoding is proposed in Figure~\ref{fig:encoding} with a running example.
Intuitively, this process is akin to building the abstract syntax tree of a program. Formally, the encoding gives 
a graph 
$\mathcal{G}(V, f, E)$ with $V = V_\textsc{var} \cup V_\textsc{cst} \cup V_\textsc{val} 
\cup V_\textsc{ope} \cup V_\textsc{mod}$ is a set containing the five types of vertices, $f = f_\textsc{var} \cup f_\textsc{cst} \cup f_\textsc{val} \cup  
f_\textsc{ope} \cup f_\textsc{mod}$ is the set of specific features attached to each vertex, and $E$ is the set of edges connecting vertices. Each type is defined as follows.

\begin{description}
\item[Values ($V_\textsc{val}$).]
A \textit{value-vertex} is introduced for each constant appearing in $\mathcal{P}$. Such values can appear inside either a domain 
or a constraint. All the values are distinct, i.e., 
they are represented by a unique vertex. The type of the value
(integer, real, etc.) is added as a feature ($f_\textsc{val}$) to each \textit{value-vertex}, using a one-hot encoding.
\item[Variables ($V_\textsc{var}$).]
A \textit{variable-vertex} is introduced for each variable appearing in $\mathcal{P}$. 
A vertex related to a variable $x \in X$ is
connected through an edge to each value $v$ inside the domain $D(x)$.
Such as the \textit{value-vertices}, all variables are represented
by a unique vertex. The type of the variable (boolean, integer, set, etc.)
is added as a feature ($f_\textsc{var}$) using a one-hot encoding.
\item[Constraints ($V_\textsc{cst}$).]
A \textit{constraint-vertex} is introduced for each constraint appearing in $\mathcal{P}$. 
A vertex related to a constraint $c \in C$ is connected through an edge to each value $v$ that is present in the constraint. The type of the constraint (\textit{inequality}, \textit{allDifferent}, \textit{table}, etc.) is added as a feature ($f_\textsc{cst}$) using a one-hot encoding.
\item[Operators ($V_\textsc{ope}$).]
Tightly related to constraints, \textit{operator-vertices} are used to unveil the combinatorial structure of constraints. 
Specifically, operators represent modifications or restrictions on variables involved in constraints (e.g.,
arithmetic operators, logical operators, views, etc.).
An \textit{operator-vertex} is added each time a variable $x \in X$ is modified by such operations. 
The vertex is connected to the vertices related to the impacted variables.
The type of the operator (\textsc{+}, \textit{$\times$}, \textit{$\land$}, \textit{$\lor$}, etc.) is added as a feature ($f_\textsc{ope}$) using a one-hot encoding. If the operator uses a numerical value to modify a variable, this value is used as a feature ($f_\textsc{ope}$) as well.
\item[Model  ($V_\textsc{mod}$).] There is only one \textit{model-vertex} per graph.
This vertex is connected to all \textit{constraint-vertices} and all \textit{variable-vertices} involved in the objective
function.
Its semantics is to gather additional information about the instance (e.g.,  the
direction of the optimization) by means of its feature vector ($f_\textsc{mod}$). 
\end{description}

As a final remark, this encoding can be used to represent any combinatorial problem instance in a unique way. To do so, a parser for each constraint, describing how a constraint has to be split with the involved operators and 
variables, must be implemented.
We currently support all the constraints formalized  in \textsc{XCSP3-core} modeling format \cite{Boussemart_Lecoutre_Audemard_Piette_2022} and used
for the mini-solver tracks of the \textsc{XCSP-2023} competition:
\textit{binary intension}, \textit{table}, \textit{negative table}, \textit{short table}, \textit{element} and \textit{sum}.
Our repository contains the required documentation to build a graph from an
instance in the \textsc{XCSP3-core} format with a description of all the features considered.

\section{Learning from the Encoding with a Graph Neural Network}
\label{sec:gnn}
To achieve the next step and \textit{learn} from this representation, we designed a tailored \textit{graph neural network} (GNN) architecture
to leverage this encoding.
A GNN is a specialized neural architecture designed to compute a latent representation (known as an \textit{embedding}) for each node of a  given  graph~\cite{Scarselli_Gori_Tsoi_Hagenbuchner_Monfardini_2009}. This process involves iteratively aggregating information from neighboring nodes. Each aggregation step is denoted as a \textit{layer} of the GNN and incorporates learnable weights. There are various ways to perform this aggregation, leading to different variants of GNNs documented in the literature \cite{li2016gated,monti2017geometric,velivckovicgraph}.
The model is differentiable and can then be trained with gradient descent methods.

Let $\mathcal{G}(V, f, E)$ be the graph encoding previously obtained, and let 
$h^{[i]}_{t,v} \in \mathbb{R}^{p}$ be a $p$-dimensional vector representation
of a vertex $v \in V_t$ ($t$ refers to a vertex type from $\mathcal{T} : \{\textsc{var},\textsc{val},\textsc{cst},\textsc{ope},\textsc{mod}\}$) at iteration $i \in \{ 0, \ldots, I\}$. 
The inference process of a GNN consists in computing the next representations ($h^{[i+1]}_{t,v}$) from the previous ones
for each vertex. This operation is commonly referred to as \textit{message passing}.
First, we set $h^{[0]}_{t,v}  =  f_{t,v}$ for each type, where 
$f_{t,v}$ is the vector of features related to vertex $v \in V_{t}$.
Then, the representations at each iteration are obtained with 
$\textsc{LayerNorm}$~\cite{ba2016layer} and LSTM layers~\cite{Hochreiter_Schmidhuber_1997}. The whole inference process is formalized in Algorithm~\ref{algo:inference}. First, the initial embedding of each vertex
is set to its feature (line~\ref{eql:1}) and the hidden states of the LSTM are initialized to 0 
(line~\ref{eql:2}), as commonly done. Then, $I$ steps of message-passing are carried out (main loop).
At each iteration $i$, a \textit{message} ($\mu^{[i]}_{t_1,v}$) is obtained for each vertex.
The computation is done in three steps (line~\ref{eql:4}):
(1) the embedding of each neighbor of a given type is summed up,
(2) the resulting value is fed to a standard multi-layer perceptron ($\mathsf{MLP}^{\mathsf{in}}_{t_1,t_2}$), note that
there is a specific module for each edge type,
(3) the messages related to each type are concatenated together ($\bigoplus$) to obtain the global message
for each vertex. Notation  $\mathcal{N}_{t_2}(v)$ refers to the set of neighbors of vertex $v$ of type $t_2$. The result is then given as input to an LSTM cell (line~\ref{eql:5}, one cell for each type) and is used
to obtain the embedding at the next layer. We note that each LSTM  has its internal state ($\gamma^{[i]}_{t,v}$) updated. 
At the end of the loop, each vertex has a specific embedding ($h^{[I]}_{t,v}$). After the last iteration, we compute the vertex-type dependent output by passing $h^{[I]}_{t,v}$ through a standard multi-layer perceptron $\mathsf{MLP}^{\mathsf{out}}_{t}$ (line~\ref{eql:6}). This produces the output embeddings for all nodes, which are then averaged (line~\ref{eql:7}). Finally, the sigmoid function ($\sigma$) is used to obtain an output between 0 and 1.

\begin{algorithm}[!ht]
$\triangleright$ \textbf{Pre:} $\mathcal{G}(V_{\textsc{var}, \textsc{cst},\textsc{val}, 
\textsc{ope},\textsc{mod}}, f_{\textsc{var}, \textsc{cst},\textsc{val}, 
\textsc{ope},\textsc{mod}}, E)$ is the graph encoding.

$\triangleright$ \textbf{Pre:} $ \mathcal{T} : \{\textsc{var},\textsc{val},\textsc{cst},\textsc{ope},\textsc{mod}\}$ is the set of \textit{vertex-types}.

$\triangleright$ \textbf{Pre:} $I$ is the number of iterations of the GNN.

~

$h^{[0]}_{t,v} := f_{t,v} ~ ~ ~      \forall  v \in  V_t, \forall  t \in  \mathcal{T}  $ \label{eql:1}

$\gamma^{[0]}_{t,v} := 0 ~ ~ ~ ~ ~  ~ ~  ~     \forall  v \in  V_t,  \forall  t \in  \mathcal{T} $  \label{eql:2}

\For{$i  ~\mathbf{from}~ 1~\mathbf{to}~ I$ }{  \label{eql:3}

$\mu^{[i]}_{t_1,v} := \bigoplus\limits_{t_2 \in \mathcal{T}}  \mathsf{MLP}^{\mathsf{in}}_{t_1,t_2}\Big( \sum\limits_{u \in \mathcal{N}_{t_2}(v)} 
h^{[i-1]}_{t_2,u}\Big)  ~ ~ ~      \forall v \in V_{t_1}, \forall t_1 \in \mathcal{T}$ \label{eql:4}

$h^{[i]}_{t,v}, \gamma^{[i]}_{t,v} := \mathsf{LSTM}_t\Big( \mu^{[i]}_{t,v}, \gamma^{[i-1]}_{t,v} \Big)  ~ ~ ~      \forall v \in V_{t} , \forall t \in \mathcal{T}$ \label{eql:5}

}

$ \nu_{t,v} :=  \mathsf{MLP}^{\mathsf{out}}_{t} \Big(h^{[I]}_{t,v} \Big) ~ ~ ~    \forall  v \in  V_t, \forall t \in \mathcal{T}$ \label{eql:6}

$\hat{y} := \sigma\Big( \frac{1}{|\mathcal{T}|\times |V|} \sum\limits_{t \in \mathcal{T}} \sum\limits_{v \in V_t} \nu_{t,v}  \Big)$ \label{eql:7}

\Return $\hat{y}$ \label{eql:8}

\caption{Inference process of our GNN architecture.}
\label{algo:inference}
\end{algorithm}

\section{Experiments}
\label{sec:exp}

This section evaluates our approach on combinatorial task, focusing on four problems: Boolean satisfiability (\textsc{SAT}), graph coloring (\textsc{COL}), knapsack (\textsc{Knap}), and the traveling salesperson problem (\textsc{TSP}) with \textsc{TSP-Ext} (table constraint) and \textsc{TSP-Elem} (element constraint) models. We trained models on the decision version of the problems, asking if a solution exists with costs under a target $k$. When there is no objective function, we have a plain constrained satisfaction problem (e.g., \textsc{SAT}). The aim is not to find the solution but to determine its existence, aligning with recent studies ~\cite{Selsam2018,Prates_Avelar_Lemos_Lamb_Vardi_2018,lemos2019graph,Liu_Zhang_Huang_Niu_Ma_Zhang_2020}. We compared our approach with problem-specific architectures and the tripartite graph of Marty et al.~(2023) ~\cite{Marty_François_Tessier_Gauthier_Rousseau_Cappart_2023}. For the latter, we extracted their graph representation and used it in our GNN. The evaluation metric considered is the accuracy in correctly predicting the answer to the decision problem. Details on experimental protocols and implementations follow.

\begin{description}
\item[Boolean Satisfiability.]
Instances are generated with the random generator of Selsam et al.~(2018)~\cite{Selsam2018}. 
Briefly, the generator builds random pairs of \textsc{SAT} instances of $n$ variables by adding new random clauses 
until the problem becomes unsatisfiable. Once the problem becomes unsatisfiable, it changes the sign of the first literal of the problem, rendering it satisfiable. On average, clauses have an arity of 8. Both the satisfiable and unsatisfiable instances are included in the dataset. 
Still following Selsam et al.~(2018), we built a dataset containing millions of instances having 10 to 40 literals. The \textsc{SAT}-dependent architecture considered in our analysis has been also introduced in the same paper. We used a training set of size 3,980,000 and a validation set of size 20,000.
\item[Traveling Salesperson Problem.]
Instances are generated with the random generator of Prates et al.~(2018)~\cite{Prates_Avelar_Lemos_Lamb_Vardi_2018}.
The generation consists in (1) creating $n$ points in a $(\sqrt{2}/2 \times \sqrt{2}/2)$ square, 
(2) building the distance matrix with the Euclidean distance, and (3) 
solving them using the Concorde solver~\cite{applegate2006concorde} to obtain the optimal tour cost $C^*$. 
Two instances are then created: a feasible one and an unfeasible one with target costs of $1.02 C^*$ and $0.98 C^*$, respectively. We build two \textsc{TSP} models:
a first one where the distance constraints are expressed with an extension constraint (\textsc{TSP-Ext}), and a second one with an element constraint (\textsc{TSP-Elem}). The motivation is to analyze the impact 
of the model on the resulting graph encoding and the performances.
Still following Prates et al.~(2018), 
we built the dataset with a number of cities $n$ sampled uniformly from $20$ to $40$.
The \textsc{TSP}-dependent architecture considered in our analysis has been also introduced in the same paper. We used a training set of size 850,000 and a validation set of size 50,000.
\item[Graph Coloring.]
Instances are generated following Lemos et al.~(2019)~\cite{lemos2019graph}. 
This generator builds graphs with 40 to 60 vertices. 
For each graph, a single edge is added such that the $k$-coloriability is altered. 
The instances are produced in pairs: one where the optimal value is $k$ and another one where it is higher. 
Our encoding leverages a standard model of the graph coloring featuring binary intension ($\neq$) constraints.
 The \textsc{Col}-dependent architecture has been also introduced in the same paper. We used a training set of size 140,000 and a validation set of size 10,000.
\item[Knapsack.]
We built instances containing 20 to 40 items and solved them to optimality to find the optimal value $V^*$.
Then, we created two instances, a feasible one and an unfeasible one with a target cost of $1.02 V^*$ and $0.98 V^*$, respectively.
Our encoding leverages a standard knapsack model featuring a \textsc{sum} constraint. The \textsc{Knapsack}-specific model was based on 
the model of Liu et al.~(2020)~\cite{Liu_Zhang_Huang_Niu_Ma_Zhang_2020}. We used a training set of size 950,000 and a validation set of size 50,000.
\item[Implementation Details.]
All models were trained with PyTorch~\cite{paszke2019pytorch} and PyTorch-Geometric~\cite{fey2019fast} on a single Nvidia V100 32 GB GPU for up to 4 days or until convergence. We selected the model having the best performance on the validation set.
To make the comparisons between specific and generic graphs fair, we trained all our models using a single GPU, and we 
tuned each model by varying the number of hidden units in the MLP and LSTM layers. 
Concerning the problem-specific architectures, we reused to same hyperparameters as described in their original publications. 
All models are trained with Adam optimizer~\cite{KingmaB14} 
coupled with a learning rate scheduler and a weight decay of $10^{-8}$. 
 The main hyperparameters used for our different models are detailed in the accompanying code.
All our models are expressed using \textsc{XCSP3} formalism. We implemented a parser to build the graph from this representation.
Our code and trained models are publicly available.
\end{description}



\subsection{Results: Accuracy of the Approaches}

Table~\ref{table:main-results} summarizes the accuracy in predicting the correct answer 
on the validation set for each problem and baseline. 
Interestingly, we observe that our approach achieves similar or close performances as the problem-specific architectures for all the problems. 
We see it as a promising result, as our approach, thanks to its genericity, can be directly used for all the problems without designing a new dedicated architecture.
On the other hand, the approach of Marty et al.~(2023)~\cite{Marty_François_Tessier_Gauthier_Rousseau_Cappart_2023}, fails to achieve similar results except for the graph coloring. This is because this representation
does not preserve the combinatorial structure of complex constraints (\textsc{Col} has only  constraints like $x_1 \neq x_2$).

\begin{table}[ht!]
\centering
\caption{Prediction accuracy on holdout test set}
\label{table:main-results}
\begin{tabular}{x{3cm}||x{1.5cm}|x{1.5cm}|x{2cm}|x{1.5cm}|x{1.5cm}} 
\toprule
\textbf{Architecture}                 & \textsc{SAT}      & \textsc{TSP-Ext}   & \textsc{TSP-Elem}   &  \textsc{Col}  & \textsc{Knap} \\ 
\midrule
\midrule
\textbf{Problem-specific}             & 94.3\%           & \multicolumn{2}{c|}{\textbf{96.3\%}}           & 77.0 \%        & \textbf{98.8\%} \\ \hline
\textbf{Tripartite~\cite{Marty_François_Tessier_Gauthier_Rousseau_Cappart_2023}}
                                      & 50.0\%              & \multicolumn{2}{c|}{50.0\%}                & \textbf{84.6\%}         & 50.0\% \\ \hline
\textbf{Ours}                         & \textbf{94.4\%}            & 84.5\%             & 91.4\%              & 84.4\%         & 97.9\% \\ 
\bottomrule
\end{tabular}

\end{table}

While \textsc{TSP-Elem} achieves results close to the TSP-specific approach, \textsc{TSP-Ext} falls short significantly. This highlights the importance of using an appropriate combinatorial model
for the encoding. Specifically, the encoding \textsc{TSP-Elem} has
a size of $1841$ vertices and $13042$ edges, while  \textsc{TSP-Ext} yields a graph of $5661$ vertices and $28322$ edges, for a same instance of 40 cities.
This is consequently larger encoding, which is not desirable as it makes the model harder to train.


\subsection{Analysis: Generalization to Larger Instances}

\begin{figure}[!ht]
    \centering
\includegraphics[width=1\linewidth]{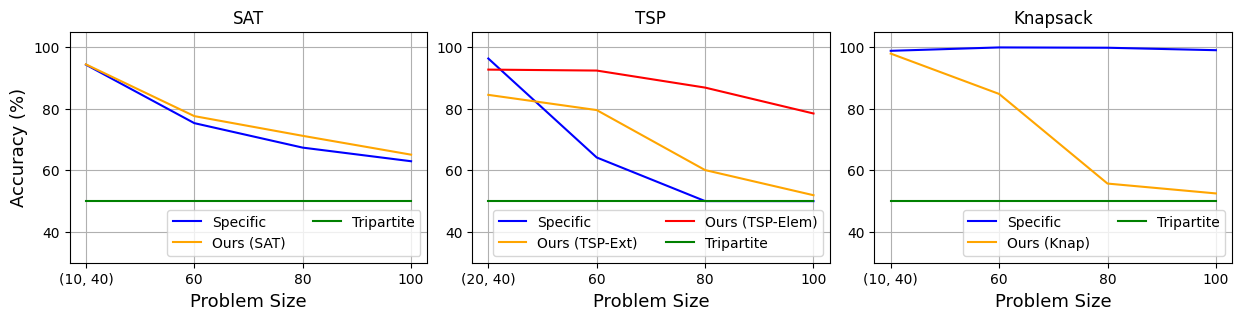}
    \caption{Analysis of the generalization ability on larger instances.}
    \label{fig:generalization}
\end{figure}

Figure~\ref{fig:generalization} shows the generalization ability of the previous model, without retraining, on new instances with 60, 80 and 100 variables (5000 instances for each size).  
We observe that our generic representation provides a better generalization than the problem-specific architecture for \textsc{SAT} and \textsc{TSP}. Notably, \textsc{TSP-Elem} offers a far better generalization than  \textsc{TSP-Ext}, confirming the impact of the input model and the size of the graph.
Interestingly, we observe a strong generalization ability of the specific architecture for the knapsack.
Our preliminary analysis indicates that this is achieved thanks to a GNN aggregation function
based on a weighted sum. Understanding in detail the root causes of this generalization is part of our future work.


\subsection{Discussion: Limitations and Challenges}
Although the empirical results show the promise of this generic architecture, some limitations must be addressed. First, the training time required to obtain such results is prohibitive, although we have considered only relatively small instances so far. This is mainly because our encoding generates large graphs. This opens the door to integrating compression methods, dedicated to
shrinking the size of the encoding without losing information. A parallel can be done with the 
\textsc{smartTable}~\cite{mairy2015smart} constraints, which encodes tables more compactly. It also
highlights the importance of having a good input model, yielding a small embedding (e.g.,\textsc{TSP-Elem} versus \textsc{TSP-Ext}).
Besides, the current task is restricted to solving the decision version of  problems with an end-to-end approach. The next step will be to integrate this architecture in a full-fledged solver, such as what has been proposed by Gasse et al.~(2019) for binary mixed-integer programs~\cite{Gasse_Chetelat_Ferroni_Charlin_Lodi_2019}, and by Cappart et al.~(2021) for constraint programming~\cite{cappart2021combining}.
Finally, the generalization ability of a model trained for a specific problem (e.g., on the \textsc{TSP}) to
a similar one (e.g., \textsc{TSP} with time windows) is an interesting aspect to investigate.

\section{Conclusion and Perspectives}

This paper introduced a first version of a generic procedure to encode arbitrary
combinatorial problem instances into a graph for learning-based approaches. 
The encoding proposed is injective, meaning that each encoding can be obtained by only one instance.
Besides, a tailored graph neural network has been proposed to learn from this encoding.
Experimental results showed that our approach could achieve similar results as problem-specific architectures,
without requiring the need to hand-craft a dedicated representation.
All the constraints involved in the 2023 mini-track competition of  \textsf{XCSP3} are currently supported.
Adding new constraints only requires the implementation of a parser.
Our next steps are to challenge the approach on bigger and more complex problems, and 
on other combinatorial tasks (e.g., learning branching heuristics).

\section*{Acknowledgement}
This research has been mainly funded thanks to a NSERC Discovery Grant (Canada) held by Quentin Cappart.
This research received funding from the European Union’s Horizon 2020 research and innovation program under grant agreement No 101070149, project Tuples.

%
%
%
\bibliographystyle{splncs04}
\bibliography{references}
\end{document}